# An Improved *k*-Nearest Neighbor Algorithm for Text Categorization[1]

Li Baoli[1], Yu Shiwen[1], and Lu Qin[2]

[1] Institute of Computational Linguistics
Department of Computer Science and Technology
Peking University, Beijing, P.R. China, 100871
E-mail: {libl, yusw}@pku.edu.cn

[2] Department of Computing,
The Hong Kong Polytechnic University
Hung Hom, Kowloon, Hong Kong
E-mail: csluqin@comp.polyu.edu.hk

**Abstract:** $k$ is the most important parameter in a text categorization system based on *k*-Nearest Neighbor algorithm (*k*NN). In the classification process, $k$ nearest documents to the test one in the training set are determined firstly. Then, the predication can be made according to the category distribution among these $k$ nearest neighbors. Generally speaking, the class distribution in the training set is uneven. Some classes may have more samples than others. Therefore, the system performance is very sensitive to the choice of the parameter $k$. And it is very likely that a fixed $k$ value will result in a bias on large categories. To deal with these problems, we propose an improved *k*NN algorithm, which uses different numbers of nearest neighbors for different categories, rather than a fixed number across all categories. More samples (nearest neighbors) will be used for deciding whether a test document should be classified to a category, which has more samples in the training set. Preliminary experiments on Chinese text categorization show that our method is less sensitive to the parameter $k$ than the traditional one, and it can properly classify documents belonging to smaller classes with a large $k$. The method is promising for some cases, where estimating the parameter $k$ via cross-validation is not allowed.

## 1  Introduction

*k*-Nearest Neighbor is one of the most popular algorithms for text categorization[1]. Many researchers have found that the *k*NN algorithm achieves very good performance in their experiments on different data sets [2][3][4].

---

The idea behind *k*-Nearest Neighbor algorithm is quite straightforward. To classify a new document, the system finds the *k* nearest neighbors among the training documents, and uses the categories of the *k* nearest neighbors to weight the category candidates [1]. One of the drawbacks of *k*NN algorithm is its efficiency, as it needs to compare a test document with all samples in the training set. In addition, the performance of this algorithm greatly depends on two factors, that is, a suitable similarity function and an appropriate value for the parameter *k*.

In this paper, we focus on the selection of the parameter *k*. In the traditional *k*NN algorithm, the value of *k* is fixed beforehand. If *k* is too large, big classes will overwhelm small ones. On the other hand, if *k* is too small, the advantage of *k*NN algorithm, which could make use of many experts, will not be exhibited. In practice, the value of *k* is usually optimized by many trials on the training and validation sets. But this method is not feasible in some cases where we have no chance to do cross-validation, such as online classification. To deal with this problem, we propose a revised *k*-Nearest Neighbor algorithm, which uses different *k* values for different classes, rather than a fixed *k* value for all classes.

In Section 2, we explain our modification of the traditional *k*NN algorithm in details. Then, we test our method on a Chinese text categorization problem in Section 3. Conclusion will be given in Section 4.

## 2  Modification of the Traditional *k*NN Algorithm

While using *k*NN algorithm, after *k* nearest neighbors are found, several strategies could be taken to predict the category of a test document based on them. But a fixed *k* value is usually used for all classes in these methods, regardless of their different distributions. Equation (1) and (2) below are two of the widely used strategies of this kind method.

$$y(d_i) = \arg\max_k \sum_{x_j \in kNN} y(x_j, c_k) \qquad (1)$$

$$y(d_i) = \arg\max_k \sum_{x_j \in kNN} Sim(d_i, x_j) y(x_j, c_k) . \qquad (2)$$

where $d_i$ is a test document, $x_j$ is one of the neighbors in the training set, $y(x_j, c_k) \in \{0,1\}$ indicates whether $x_j$ belongs to class $c_k$, and $Sim(d_i, x_j)$ is the similarity function for $d_i$ and $x_j$. Equation (1) means that the predication will be the class that has the largest number of members in the *k* nearest neighbors; whereas equation (2) means the class with maximal sum of similarity will be the winner. The latter is thought to be better than the former and used more widely.

In general, the document distribution of different classes in the training set is uneven. Some classes may have more samples than others. Therefore, it is very likely that a fixed *k* value will result in a bias on large classes. For example, when using the strategy indicated by equation (2), many tiny similarity values would accumulate to a large one, which may improperly make a large class the final decision. To overcome this problem, we propose a different strategy as follows.

When we get the original *k* nearest neighbors, we compute the probability that one document belongs to a class by using only some top *n* nearest neighbors for that class, where *n* is derived from *k* according to the size of a class $c_m$ in the training set. In other words, we use different numbers of nearest neighbors for different classes in our method. For larger classes, we use more nearest neighbors. The dynamic selection is based on the class distribution in the training set. To make the comparison between classes reasonable, we derive the probabilities from the proportion of the similarity sum of neighbors belonging to a class to the total sum of similarities of all selected neighbors for that class. Equation (3) gives the decision function in our improved *k*NN algorithm.

$$y(d_i) = \arg\max_m \frac{\sum_{x_j \in top\_n\_kNN(c_m)} Sim(d_i, x_j) y(x_j, c_m)}{\sum_{x_j \in top\_n\_kNN(c_m)} Sim(d_i, x_j)} . \quad (3)$$

where,

$$top\_n\_kNN(c_m) = \{ top\ n\ \text{neighbors in the original k nearest neighbors kNN} \mid n = \left\lceil \frac{k \times N(c_m)}{\max\{N(c_j) \mid j = 1..Nc\}} \right\rceil \}.$$

Note that $N(c_m)$ denotes the size of the class $c_m$ in the training set, and $\max\{N(c_j)|j=1..Nc\}$ is the size of the largest class in the same set.

## 3 Experiments and Discussion

### 3.1 Corpus and Setting

The corpus we used is provided by the Computer Network and Distributed Systems Laboratory[2], Department of Computer Science and Technology, Peking University, which contains 19,892 Chinese web pages with a total of 278M bytes. There are 12 top-categories and 1,006 sub-categories in the corpus. We only used the 12 top-categories and in a one-label setting (i.e., a document will be assigned only one class label). Table 1 lists the 12 top-categories and their distributions in the corpus.

**Table 1.** 12 top-categories and their distributions in the corpus

| No. | Category | Number | Percentage |
|---|---|---|---|
| 1 | Humanities and Arts | 764 | 3.84 |
| 2 | News and Media | 449 | 2.26 |
| 3 | Business and Economy | 1642 | 8.25 |
| 4 | Entertainment | 2637 | 13.26 |
| 5 | Computer and Internet | 1412 | 7.10 |
| 6 | Education | 448 | 2.45 |
| 7 | Region and Organization | 1343 | 6.75 |
| 8 | Science | 2683 | 13.49 |
| 9 | Government and Politics | 493 | 2.48 |
| 10 | Social Science | 2763 | 13.89 |
| 11 | Health | 3180 | 15.99 |
| 12 | Society and Culture | 2038 | 10.25 |

---

[2] http://net.cs.pku.edu.cn

We evenly divided the corpus into 10 parts at random. In the following experiments, we use two of the ten sub-corpora: one for training, and the other for test.

In our experiments, a document is represented by a space vector, the dimensions of which correspond to Chinese words. The HTML tags were ignored here. We derived Chinese words from the raw text by using the word segmentation package SEGTAG[3]. And we used the term weighting scheme given in equation (4) (the same as the standard representation "ltc" [1]). The cosine function is used to compute the similarity between two documents.

$$x_i = \frac{(1+\log(TF(w_i,d))) \cdot \log(\frac{|D|}{DF(w_i)})}{\sqrt{\sum_j [(1+\log(TF(w_j,d))) \cdot \log(\frac{|D|}{DF(w_j)})]^2}}. \quad (4)$$

We test 12 different values for parameter $k$, which spans from 5 to 60 with interval 5.

### 3.2 Measuring Performance

To evaluate the effectiveness of category assignments by classifiers to documents, the standard precision, recall, and $F_1$ measure are used here. Precision is defined to be the ratio of correct assignments by the system divided by the total number of the system's assignments. Recall is the ratio of correct assignments by the system divided by the total number of correct assignments. The $F_1$ measure combines precision ($p$) and recall ($r$) with an equal weight in the following form:

$$F_1(p,r) = \frac{2rp}{r+p}. \quad (5)$$

These scores can be computed for the binary decisions on each individual category first and then be averaged over categories. Or, they can be computed globally over all the $n*m$ binary decisions where $n$ is the number of total test documents, and $m$ is the number of categories in consideration. The former way is called macro-averaging and the latter micro-averaging. It is understood that the micro-averaged scores (recall, precision, and $F_1$) tend to be dominated by the classifier's performance on common categories, and that the macro-averaged scores are more influenced by the performance on rare categories.

### 3.3 Results and Discussion

Table 2 gives the experimental results of two $k$NN algorithms with different $k$ values. $k$NN-A represents the traditional one, and $k$NN-B denotes our modified version. Figures in bold style indicate that they are the highest ones in each column, while figures in italic style with underscore are the lowest ones.

$k$NN-A achieved its best performance when $k$ is around 10, but $k$NN-B performed well at 15. As $k$ increases, the performance of $k$NN-A declines more rapidly than that of $k$NN-B. When $k$ reachs 60, the macro-averaging $F_1$ measure of algorithm $k$NN-A decreases by 13.32%, from 68.11% to 59.04%, and the micro-averaging $F_1$ measure changes from 72.24% to 65.93%. On average, the performance of $k$NN-B is at least 1.4% higher than that of $k$NN-A algorithm. Relative lower standard deviations of algorithm $k$NN-B suggest that our modified $k$NN algorithm behaves quite stably with different values of $k$.

---
[3] http://www.icl.pku.edu.cn/nlp-tools/segtagtest.htm

**Table 2**. Performances of two $k$NN algorithms with different $k$ (Unit: %)

| $k$ | $k$NN-A | | | | $k$NN-B | | | |
| --- | --- | --- | --- | --- | --- | --- | --- | --- |
| | Micro-Avg. | Macro-Avg. | | | Micro-Avg. | Macro-Avg. | | |
| | Pre=Rec=$F_1$ | Pre | Rec | $F_1$ | Pre=Rec=$F_1$ | Pre | Rec | $F_1$ |
| 5 | **72.35** | *70.74* | **67.24** | 67.96 | 71.69 | *67.10* | 66.57 | 66.02 |
| 10 | 72.24 | 73.17 | 66.49 | **68.11** | 71.69 | 67.88 | **67.56** | 66.29 |
| 15 | 71.44 | 74.40 | 64.81 | 67.06 | **71.94** | 69.53 | 67.15 | **67.05** |
| 20 | 70.17 | 73.75 | 63.34 | 65.60 | 71.08 | 70.58 | 66.39 | 66.75 |
| 25 | 69.26 | 74.15 | 61.66 | 64.17 | 70.68 | 70.68 | 65.75 | 66.18 |
| 30 | 68.10 | 73.37 | 60.03 | 62.54 | 70.73 | **72.48** | 65.90 | 66.92 |
| 35 | 67.54 | 74.31 | 59.34 | 62.06 | 69.87 | 71.55 | 64.95 | 65.76 |
| 40 | 67.04 | 74.53 | 57.98 | 60.90 | 69.51 | 71.38 | 64.36 | 65.29 |
| 45 | 66.89 | 74.84 | 57.21 | 60.31 | 68.66 | 70.72 | 63.21 | 64.19 |
| 50 | 66.43 | 74.17 | 56.47 | 59.38 | 68.55 | 71.73 | 62.86 | 64.19 |
| 55 | 66.28 | 75.12 | 56.26 | 59.23 | 68.40 | 71.93 | 62.65 | 64.13 |
| 60 | *65.93* | **75.86** | *55.94* | *59.04* | *67.75* | 71.35 | *61.72* | *63.33* |
| Avg. | 68.64 | 74.03 | 60.56 | 63.03 | 70.05 | 70.58 | 64.92 | 65.51 |
| STDev | 2.38 | 1.27 | 4.08 | 3.45 | 1.46 | 1.64 | 1.94 | 1.26 |

Figure 1 and 2 show the variation of $F_1$ measure in each top-category when using the two algorithms with different $k$, respectively. The Health class (no. 11) has quite stable $F_1$ measures in both algorithms. When $k$ increases, it only declines slightly. This may be due to larger training samples of this class, whose amount of example documents (about 318) is much more than the selected $k$s. Consider the three top-categories with fewer samples: News and Media (no. 2), Education (no. 6), and Government and Politics (no. 9). In Figure 1, the $F_1$ measures for these classes drop dramatically, especially for class 9. On the contrary, $k$NN-B algorithm achieves much better performance on smaller categories, which is observed from Figure 2. For class 6 and 9, the curves even show a tendency of rising as $k$ increases.

It is interesting to note that in both figures the $F_1$ measures of class 12 tend to drop drastically while $k$ increases. Class 10 has similar curves. We found that these two classes are very close to each other thus the algorithm can mix them up easily even though both of them have enough samples in the training set. In addition, these two classes are likely to be misclassified when using more nearest neighbors. Our preliminary examination suggests that the problem has more to do with the classification hierarchy. Perhaps, the class hierarchy should be modified, or we can use 1NN to do the classification for these classes.

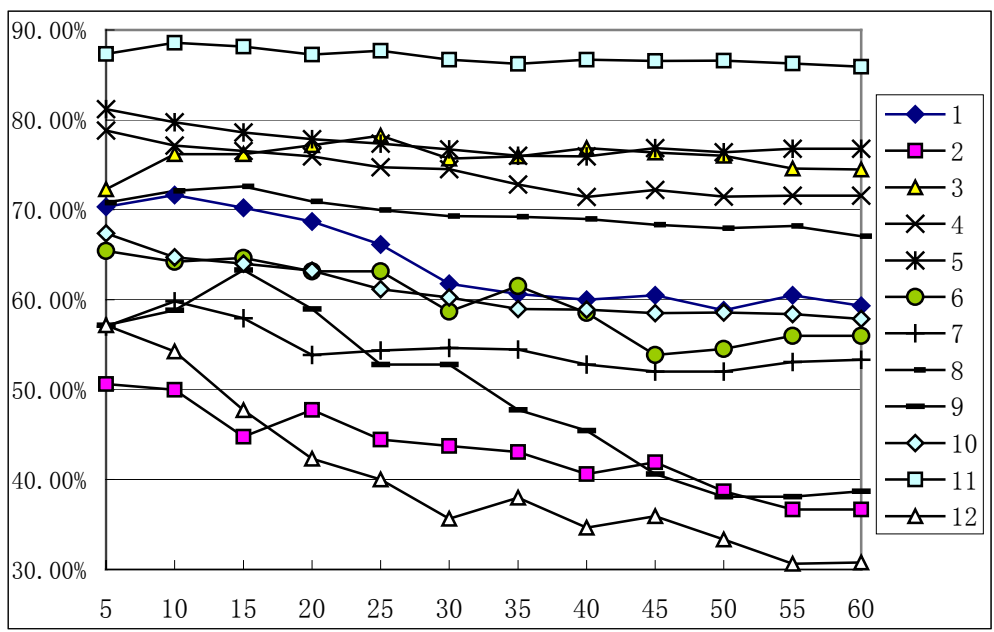

**Figure 1.** $F_1$ measures of 12 top-categories for $k$NN-A with different $k$.

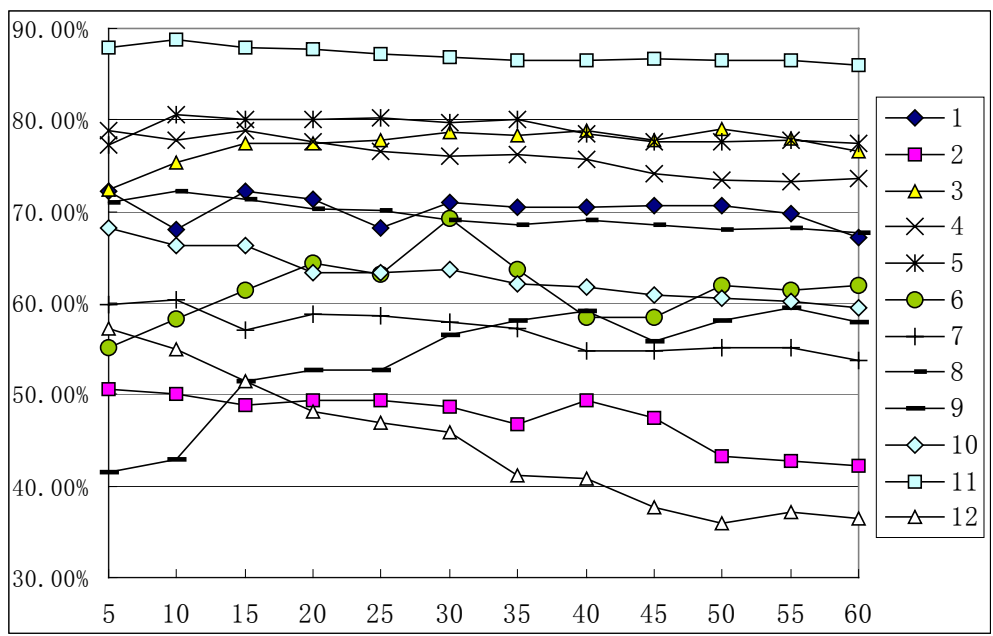

**Figure 2.** $F_1$ measures of 12 top-categories for $k$NN-B with different $k$.

We have also successfully applied this improved $k$NN algorithm in Topic Detection [5], where the system must detect new topics as the incoming stories are processed and then proceed to associate input stories with those topics online [6].

# 4   Conclusion

To alleviate the bias on larger class in traditional $k$NN algorithm, we have proposed a modified $k$NN method. For different classes, according to their distribution in the training set, we use a suitable number of nearest neighbors to predict the class of a test document. Preliminary experiments on Chinese text categorization show that our method is less sensitive to parameter $k$ than the traditional one, and it can properly classify documents belonging to smaller classes with a large $k$. The method is promising for some special cases, where estimating the parameter $k$ via cross-validation is not allowed. We plan to experiment our improved method on more different data sets in the future.

Even though we have tested the method only on Chinese text, the method should be universally applicable to classification problems for data in other languages.

# Acknowledgements

Many thanks to the corpus provider, the Computer Network and Distributed Systems Laboratory, Department of Computer Science and Technology, Peking University. This research was supported by National Natural Science Foundation of China (69973005 and 60173005). The work was also partially funded by a research grant from the Hong Kong Polytechnic University under the project code A-P203.